\documentclass{article}

\usepackage[accepted]{icml2025}

\usepackage{microtype}
\usepackage{graphicx} \graphicspath{{figures/}}
\usepackage{subfigure}
\usepackage{booktabs} 
\usepackage{tcolorbox}

\usepackage{hyperref}

\usepackage{multirow}
\usepackage{amssymb}
\usepackage{xcolor}
\usepackage{pifont} 
\newcommand{\greencheck}{{\color{green}\ding{51}}} 
\newcommand{\redx}{{\color{red}\ding{55}}}        

\usepackage{amsmath}
\usepackage{amssymb}
\usepackage{mathtools}
\usepackage{amsthm}
\usepackage{array} 

\usepackage[capitalize,noabbrev]{cleveref}

\theoremstyle{plain}

\theoremstyle{definition}

\theoremstyle{remark}

\usepackage[textsize=tiny]{todonotes}

\icmltitlerunning{Building LLM Agents by Incorporating Insights from Computer Systems}

\begin{document}

\twocolumn[
\icmltitle{Building LLM Agents by Incorporating Insights from Computer Systems}




\icmlsetsymbol{equal}{*}

\begin{icmlauthorlist}
\icmlauthor{Yapeng Mi}{bigai,hit}
\icmlauthor{Zhi Gao}{bigai,pku}
\icmlauthor{Xiaojian Ma}{bigai}
\icmlauthor{Qing Li}{bigai}
\end{icmlauthorlist}

\icmlaffiliation{bigai}{State Key Laboratory of General Artificial Intelligence, BIGAI}
\icmlaffiliation{hit}{Harbin Institute of Technology}
\icmlaffiliation{pku}{School of Intelligence Science and Technology, Peking University}

\icmlcorrespondingauthor{Qing Li}{dylan.liqing@gmail.com}

\icmlkeywords{Machine Learning, ICML}

\vskip 0.3in
]



\printAffiliationsAndNotice{}  

\begin{abstract}
LLM-driven autonomous agents have emerged as a promising direction in recent years. However, many of these LLM agents are designed empirically or based on intuition, often lacking systematic design principles, which results in diverse agent structures with limited generality and scalability. In this paper, we advocate for building LLM agents by incorporating insights from computer systems. Inspired by the von Neumann architecture, we propose a structured framework for LLM agentic systems, emphasizing modular design and universal principles. Specifically, this paper first provides a comprehensive review of LLM agents from the computer system perspective, then identifies key challenges and future directions inspired by computer system design, and finally explores the learning mechanisms for LLM agents beyond the computer system. The insights gained from this comparative analysis offer a foundation for systematic LLM agent design and advancement.
\end{abstract}  

\section{Introduction}
In recent years, autonomous agents have emerged as a promising direction, with widespread applications in various domains, such as computer use~\cite{hong2024cogagent,lin2024showui}, code assistance~\cite{wang2024openhands,yang2024swe}, and others~\cite{kim2024MDagents,fan2025videoagent}. These agents, powered by large language models (LLMs)~\cite{achiam2023gpt4} as controllers, integrate key components such as memory, tools, and action while allowing effective environmental interaction. Currently, many studies~\cite{xie2024VLMAgentSurvey,durante2024AgentAI} rely on intuition or experience to design agents, often lacking systematic design principles. As a result, the significant divergence among these designs makes it difficult to establish a standardized structure with generality and scalability. At the same time, the lack of a systematic approach to developing LLM agents limits the ability to guide future research directions. Therefore, we are faced with a critical question: \textit{How can general LLM agents be systematically constructed and evolved?}

\begin{figure*}[t]
    \centering
    \includegraphics[width=1.0\textwidth]{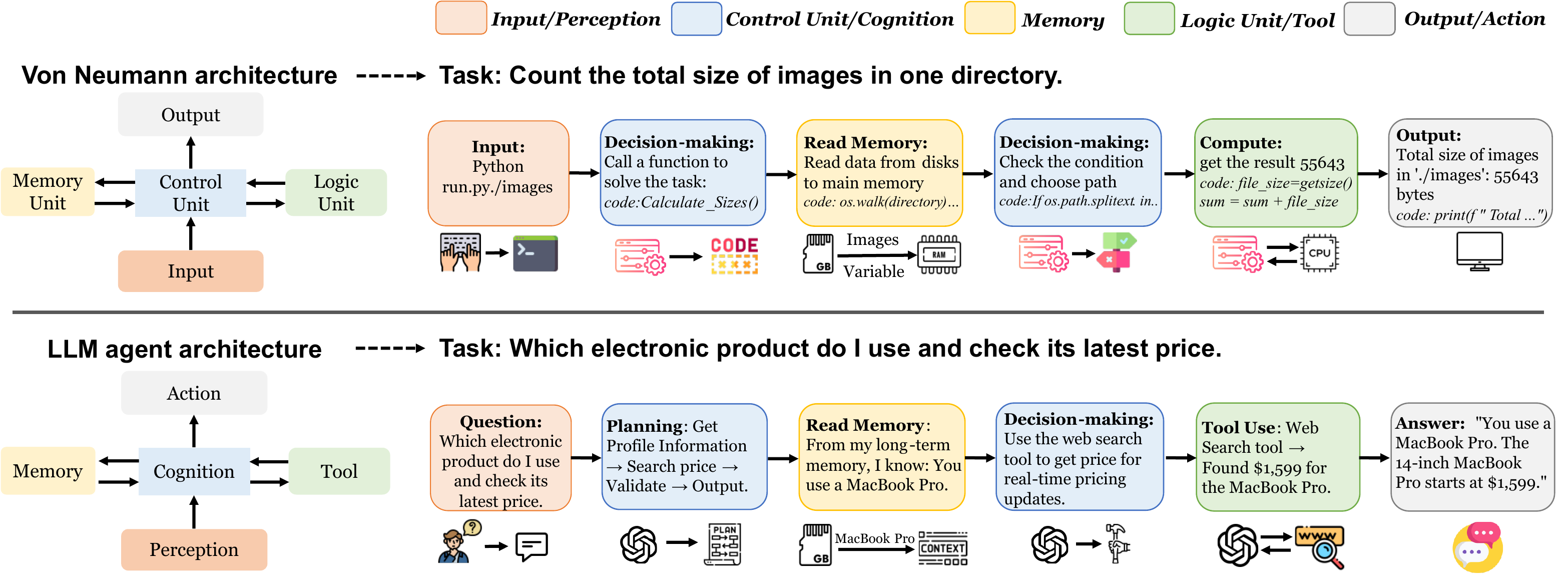}
    \caption{An analogy between von Neumann architecture and LLM agent architecture with task execution workflows (\cref{Appdenix:A}).}
    \label{fig:teaser}
\end{figure*}

In this paper, we advocate for \textbf{building LLM agents by incorporating insights from computer systems}.  Looking back at the evolution of computer system design\cite{campbell1996computer}, we find that it once witnessed a proliferation of various structural approaches. In the early days of computing, systems were often designed with specific architectures tailored to particular tasks, such as the ABC computer in 1942 for solving linear equations~\cite{vonNeumann1946}. This specificity meant that these systems lacked general-purpose capabilities. As time goes on, modern computer systems have largely converged on a unified theoretical framework: the von Neumann architecture, which has significantly advanced the field. Similarly, the design of agents today revolves around system-level considerations, much like computer systems design. A straightforward comparison between the von Neumann architecture and the agent architecture shows their high similarity in both structural design and task workflows, as shown in \cref{fig:teaser}. This parallel offers numerous opportunities for comparative analysis between these two fields. For example, both systems have memory modules, and the computer storage hierarchy can guide agents' finer-grained memory design. Furthermore, computer systems have developed several golden insights over time, such as parallelization. These principles can be effectively transferred to the design of LLM agents, providing foundational guidance for their construction.

Therefore, this paper proposes building and evolving general LLM agents by drawing insights from computer systems, including von Neumann architecture and other related principles. 
We will conduct a comprehensive analysis of the LLM agent compared with the computer systems, then investigate the potential future research direction inspired by the computer systems, and explore the learning capability of agents beyond the computer systems. 
Specifically, this work provides a comparative analysis of LLM agents for the framework, inspired by the von Neumann architecture (\cref{sec:section2}). We propose LLM agents as a collection of distinct modules that dynamically interact with the environment. This framework effectively encapsulates existing research while inspiring future directions. Then, this paper will discuss potential directions inspired by computers, including finer-grained memory, parallelization, and other potential advances(\cref{sec:section3}). Finally, we analyze current learning mechanisms and explore how such agents can learn from their environment, through which the agents could go beyond some limitations of computer systems(\cref{sec:section4}). We believe that better learning methods are key to the evolution of LLM agents. To the best of our knowledge, this is the first analogy drawn between computer systems and LLM agents, and we hope it can benefit the community.

\section{LLM Agent Framework Inspired by Von Neumann Architecture}
\label{sec:section2}
\begin{figure*}[t]
    \centering
    \includegraphics[width=0.8\textwidth]{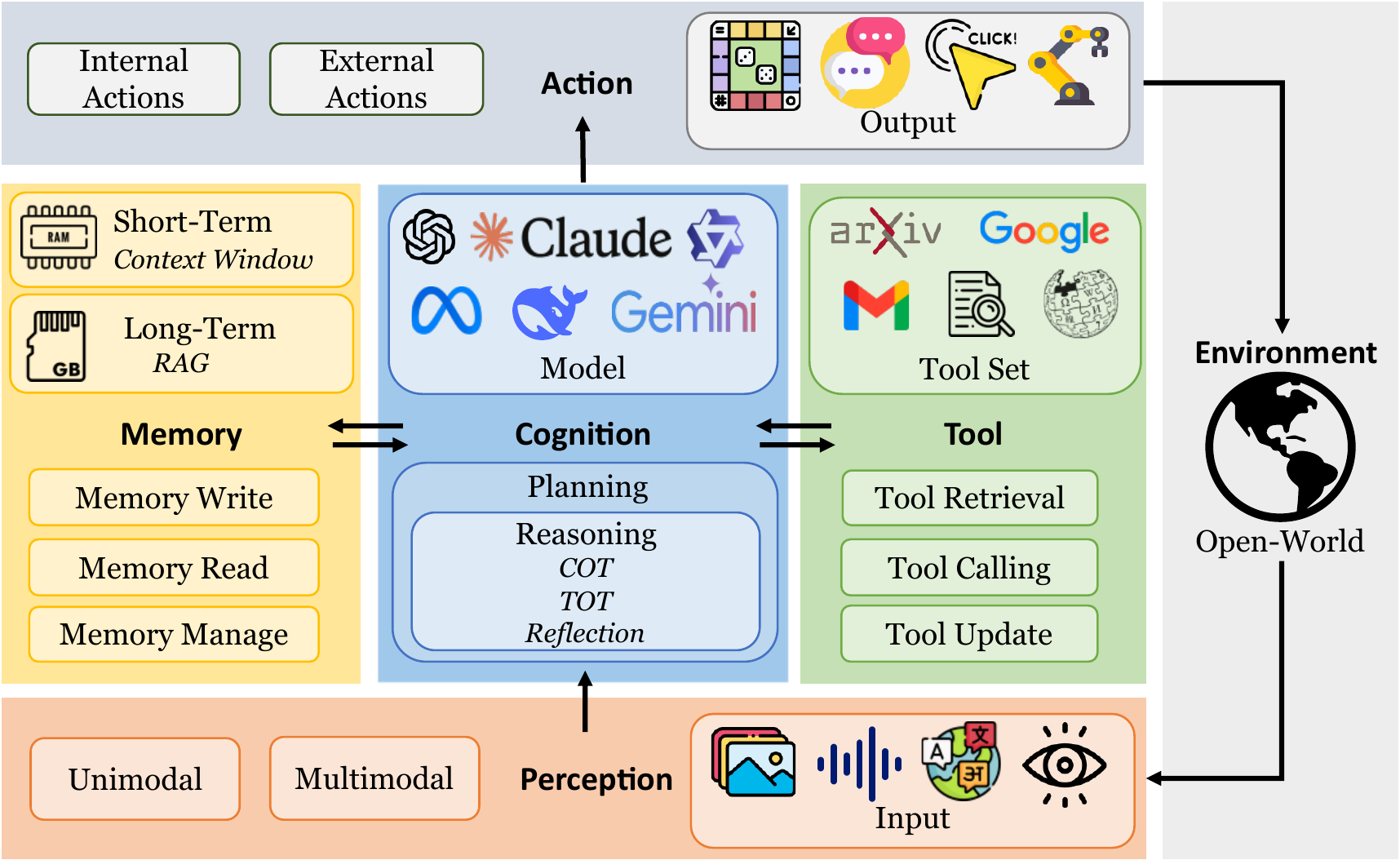} 
    \caption{A framework diagram of LLM agents inspired by the von Neumann architecture. This framework illustrates the different components of LLM agents and specific classifications within each module, along with the interactions between the modules.}
    \label{fig:main}
\end{figure*}

One important insight from computer systems is the von Neumann architecture, which established the foundational framework of modern computing by integrating a central processing unit, memory, and input/output systems~\cite{vonNeumann1946}. In this section, we follow a systematic perspective and draw inspiration from von Neumann architecture (\cref{fig:teaser}) to explore how to construct general LLM agents. We posit that an LLM agent comprises interconnected components: perception, cognition, memory, tools, and action, which dynamically interact with each other and the environment, as shown in \cref{fig:main}. We conducted a literature review based on this definition. During this process, we also conducted a comparative analysis with the von Neumann architecture, revealing shared principles in system design between the two.

To further support the foundational framework, this paper provides a concise formulation. To formalize, each module is denoted by its uppercase initial (e.g., \(P\) for perception). The framework is defined as \(F = (P, C, M, T, A) \). Specifically, the LLM agent also receives observations from the external environment, denoted \(o\). At time step \( t \), the action \( a_t \) can be derived using the following equation:
\begin{equation}
a_t = A\biggl( C\bigl( P(o_1, a_1, \dots, o_{t-1}, a_{t-1},o_t), M_{r}, T_{c} \bigr)\biggr).
\end{equation}
Here, \(M_{r}\)represents the content retrieved by the memory module, and \(T_{c}\) represents the output of the calling tool. This equation briefly illustrates how our framework generates the action \(a_{t}\) by integrating the historical sequence starting from the initial prompt-guided observation \(o_1\), followed by actions and observations, combined with memory content 
and tool outputs. More details can be found in \cref{Appdenix:B}.

\subsection{Perception}
For agents interacting in real-world environments, the capability of perception is indispensable, similar to the role of input modules in computer systems. Just as computers use input such as keyboards and cameras, agents gather various types of information, such as text, images, video, and audio. Based on existing research, the perception of agents can be divided into two parts: unimodal and multimodal.

\textbf{Unimodal}: Unimodal perception means that the agent perceives
only one type of information. Due to the significant achievements of LLMs in natural language processing, early agents used LLMs as core processors, relying on the model's inherent perception capabilities. In previous studies~\cite{wang2023voyager,shen2024hugginggpt}, these agents only utilize textual information in environments for processing and planning, and restrict accurate understanding of multimodal information, driving the development of subsequent multimodal perception agents.

\textbf{Multimodal}: With recent advancements in multimodal LLMs, more agents~\cite{hong2024cogagent,kim2024MDagents} now adopt multimodal models as controllers, enabling the integration of diverse modalities such as text, images, and videos. For complex scenarios, external encoders~\cite{fan2022MineCLIP} can also be used to convert information into forms perceivable by agents. In summary, multimodal has become a mainstream direction for agent perception.

Notably, \textbf{a crucial principle of von Neumann architecture is that various types of information are transformed into a common representation within the system}. Computer systems represent information in a data space of 0s and 1s, modern agents represent perceived information in the language space, which offers several distinct advantages. Firstly, language space demonstrates higher fault tolerance during agent task execution due to the superior generalization capabilities of natural language. Secondly, language space enables the expression of more abstract and ambiguous concepts, which is often beyond the capacity of conventional computer systems. Lastly, since the output in the process of task completion can be observed in natural language, representing information in the language space inherently provides better Interpretability.

\subsection{Cognition}
The cognition module plays a crucial role in LLM agents, much like the position of the control unit in the computer architecture. Cognition governs the agent's decisions, encompassing planning and reasoning capabilities. Planning is an essential capability for almost every LLM agent, while reasoning involves various thought processes such as chain of thought and reflection to facilitate better planning. Analogously, planning can be seen as the central function of the control unit, while reasoning represents the underlying logic that drives its decision-making processes.

\textbf{Planning}: Planning is a foundational capability of cognition. Given a goal,OK, planning decides on a sequence of actions $(a_0, a_1, \ldots, a_n)$ that will lead to a state achieving the goal. Early studies~\cite{liu2023llm+p,yao2020CALM} explored the use of external planners to assist planning. With the further enhancement of LLMs in reasoning~\cite{wei2022COT}, planning is expected to be incorporated into the controllers themselves. For example, WebDreamer~\cite{gu2024WebDreamer} uses LLMs to simulate candidate actions and evaluate their outcomes to select the optimal step. Therefore, we can see that future planning methods are likely to be centered around LLMs, requiring less scaffolding. This is largely attributed to the improvement of reasoning capabilities of LLMs, which is crucial in cognition.

\textbf{Reasoning}: Reasoning is a fundamental unit of cognition, and it supports the thinking capability of LLM agents. Objectively, the development of reasoning is closely tied to LLMs, significantly expanding their ability to think and enabling them to solve more complex problems. In recent years, the advancement of reasoning based on LLMs began with CoT~\cite{wei2022COT}, which allows the model to decompose complex problems into a sequence of intermediate reasoning steps. CoT-SC~\cite{wang2022COT-SC} builds upon CoT by generating multiple answers simultaneously and selecting the optimal one.  ToT~\cite{yao2024TOT} adopts a tree structure to explore multiple reasoning paths and self-evaluate for a globally optimal decision. These methods enable the cognition module to engage in deep thinking, thereby significantly enhancing its decision-making capabilities.

Another key capability in reasoning is known as self-refinement or reflection. This ability allows LLM agents to reflect on their past actions and further optimize future decisions. Essentially, it transforms a linear thought process into a cyclical one by treating the decision-making process as continuous. The ReAct~\cite{yao2022react} framework is a notable pioneer in this area. It enhances LLMs by integrating task-specific actions for environmental interaction and natural language reasoning for situational reflection. Reflexion~\cite{shinn2024reflexion} converts binary or scalar feedback from the environment into textual summaries, which are then incorporated as additional context for LLM agents in subsequent iterations. Most self-refinement mechanisms enable agents to improve through trial-and-error processes, enhancing their robustness.

Notably, in the history of electronic computing, the evolution of CPUs, following the unified von Neumann architecture, has consistently been the most critical development in computers. From this perspective, the cognition module of LLM agents is likely to follow a similar trajectory. In the future, advanced planning and reasoning units may emerge as the core direction for LLM agents, further underscoring the central role of cognition in their development.

\subsection{Memory}
\begin{figure}[htbp]
    \centering
    \includegraphics[width=0.48\textwidth]{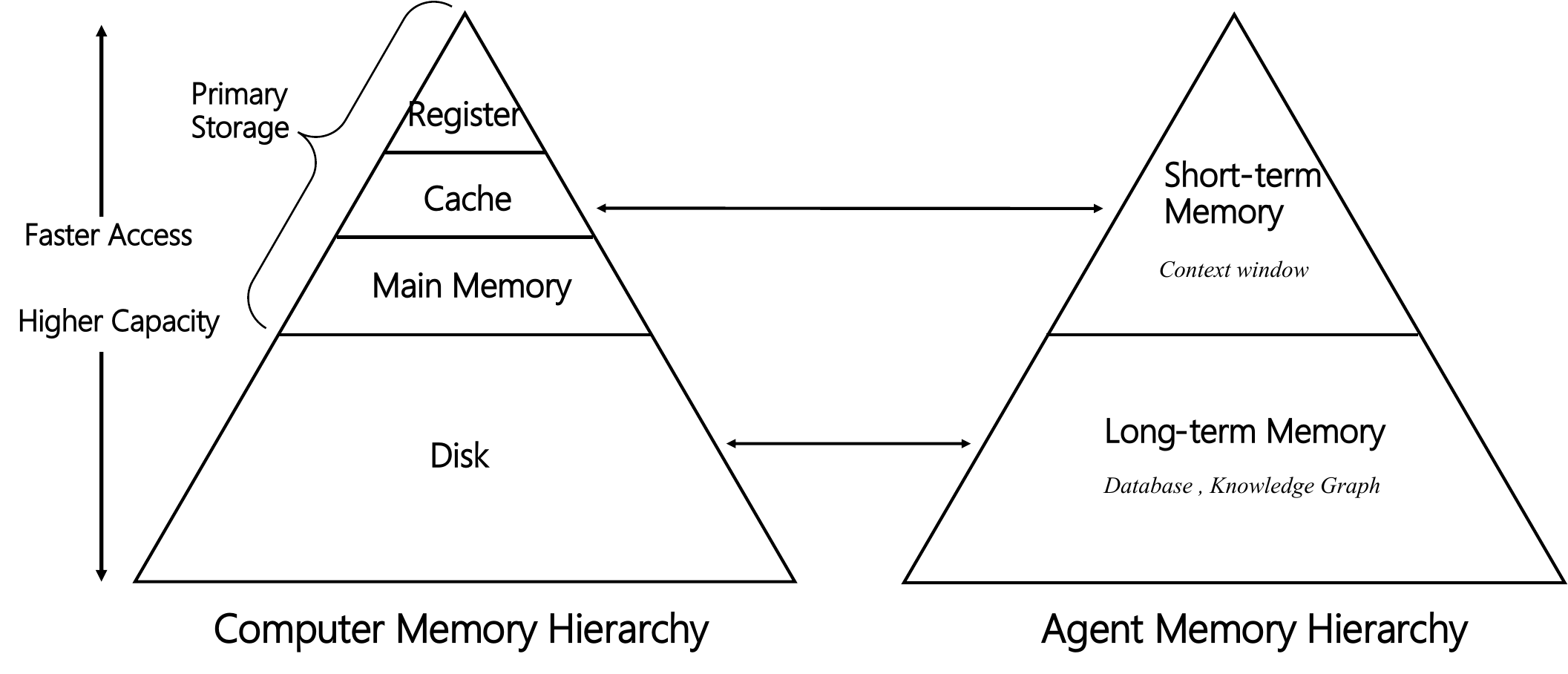}
    \caption{An intuitive comparison between computer memory hierarchy and agent memory hierarchy.}
    \label{fig:hierarchy}
\end{figure}

In the von Neumann architecture, the memory unit serves as the central repository for both data and instructions. This essential unit can be analogously extended to LLM agents, where a memory-like component is constructed for storing states, knowledge, or learned behaviors, facilitating decision-making and dynamic adaptation. In computer systems, the memory units follow a memory hierarchy, which is a good insight from computer systems. We find a strong analogy between the memory between computers and LLM agents, as illustrated in \cref{fig:hierarchy}. Short-term memory is smaller than long-term memory but is fast, and long-term memory restores information longer but short-term memory cannot. Therefore, memory of LLM agents can be categorized into short-term and long-term memory, accompanied by associated read and write operations, aligning with the aforementioned perspective.

\textbf{Short-Term Memory}: Short-term memory is often incorporated into LLMs, like primary memory in computer architecture, as illustrated in \cref{fig:hierarchy}. Short-term memory refers to an LLM agent's contextual understanding within its context window, which is critical for immediate operations. In certain agent task interaction workflows, short-term memory is often used to store contextual input-output information related to the task, including reasoning details, tool usage data, and past state information. Therefore, it is often referred to as working memory, which supports contextually appropriate responses during ongoing interactions.

\textbf{Long-Term Memory}: In contrast, long-term memory is designed to store information for extended periods, enabling LLM agents to retrieve and utilize past knowledge or skills across different interactions, like disks in the memory hierarchy (see \cref{fig:hierarchy}). Long-term memory encompasses a diverse range of content types, including text, images, codes, trajectories, profile information, and successful demonstrations, significantly broadening the scope of information that agents can leverage. Long-term memory is typically implemented through external memory systems, enhancing an agent's reliability.

\textbf{Memory Read and Write}: Memory reading and writing are fundamental operations in the von Neumann architecture, serving as the critical link between the CPU and memory. Similarly, LLM agents are increasingly equipped with mechanisms to read and write their memory. These capabilities allow agents to adapt dynamically to changing environments by editing stored information, ensuring that the memory remains relevant to specific tasks. For memory reading, most methods select memory values that exhibit the highest similarity to the query object. Furthermore, many agents adopt Retrieval-Augmented Generation~\cite{lewis2020RAG} to form a complete reading and generation workflow. Generative Agents~\cite{park2023GenerativeAgents} retrieve necessary information in text form based on relevance, recency, and importance. Memory writing often follows the storage format of memory reading. For example, in MemoChat~\cite{lu2023memochat}, agents summarize each conversation segment by identifying the main topics discussed and storing them as keys to index memory pieces. This structured approach facilitates efficient memory retrieval during reading operations. Memory reading and writing operations are collaborative, collectively enhancing the capabilities of LLM agents. Moreover, from a computational perspective, the speed of memory read and write operations remains an area to be explored for LLM agents, as it directly impacts information processing and task execution efficiency.

\subsection{Tool}
A unique feature is that LLM agents could utilize external tools, which is similar to a logic unit in computer systems. As the logic unit is the true executor of arithmetic operations, the tools within LLM agents concrete execution of tasks. Tool usage is often presented as APIs, enabling agents to select the appropriate tools based on their objectives quickly. Commonly used tools in existing methods are diverse and depend on the design of the toolset, such as calculators for arithmetic operations, and Google search for retrieving the latest information. There are two main interactions in tool use, which are named tool retrieval and tool calling. 

\textbf{Tool Retrieval}: When given a question that needs tools to solve, LLM agents perform reasoning and planning to generate responses containing tool information. Tool retrieval means selecting the right tools using a retriever or LLM. For instance, Gorilla~\cite{patil2023gorilla} employs BM25 and GPT-Index to construct a retriever to implement tool retrieval. ToolLLM~\cite{qin2023toolllm} trains a Sentence-BERT model to serve as a tool retriever, allowing highly efficient retrieval of relevant tools. 

\textbf{Tool Calling}: As for tool calling, it means a right tool is correctly provided with required parameters and executed successfully to return results for LLM agents, making it a successful call. In EasyTool~\cite{yuan2024easytool}, it improves LLMs' understanding of tool functions and parameter requirements by prompting ChatGPT to rewrite tool descriptions. ConAgents~\cite{shi2024ConAgents} presents a multi-agent collaborative framework, incorporating a dedicated execution agent responsible for parameter extraction and tool invocation. Notably, LLMs sometimes act as tools, representing the execution end beyond decision-making, not limited to specific instruments.

\subsection{Action}
Similar to the output modules in computer systems, the action module in LLM agents serves as the interface for interaction with environments. The functionality and mechanisms of the action module are typically goal-oriented and closely tied to LLM agents' operating environments. Fundamentally, the action module transforms high-level actions into low-level actions through appropriate conversions. The action module can be simply categorized into two types: internal actions and external actions.

\textbf{Internal Actions}: Internal Actions refer to the internal operations performed by LLM agents during task execution, including memory read/write, tool invocation, and reasoning actions. These internal actions modify agents' internal state, enabling it to better learn and adapt to environments. Specifically, reasoning represents a novel internal action that does not exist in traditional computer systems, highlighting that LLM agents have a broader action space due to their ability to operate within abstract semantic spaces.

\textbf{External Actions}: External actions refer to the actions output by LLM agents that target external environments. These actions are closely tied to the external context and primarily include execution actions in GUI or embodied environments, as well as interaction actions in natural language dialogue. For example, SayCan~\cite{ahn2022SayCan} generates actionable steps for robots, ShowUI~\cite{lin2024showui} outputs actions and coordinates for virtual environments, and MDagents~\cite{kim2024MDagents} perform conversational actions to provide medical advice. It is worth noting that some action formats require specific agents to use an additional action processing function to generate executable actions.

\subsection{Environment}
In computer systems, computers interact with environments in ways designed by humans. Similarly, an agent system only exerts its unique capabilities through environmental interaction. The interaction environment of agent LLMs is highly diverse; it could be a human~\cite{chen2023llava-interactive}, a game~\cite{wang2023voyager}, or a real-world setting~\cite{ahn2022SayCan}. Typically, an agent acts and receives new observations and feedback from the environment after execution, and this process is called a complete interaction loop. What's more, people design effective human-computer interaction ecosystems, such as VSCode. Similarly, there is a significant unexplored ecosystem space between LLM agents and external environments. For example, SWE-agent~\cite{yang2024swe} constructs a friendly interface between LLM agents and software engineering. This analogy highlights the potential for enhanced environmental interaction.

\begin{table*}[t]
\centering
\caption{Summary of module inclusion in landmark LLM agent studies based on the proposed framework. Specifically, we also include a summary of the learning mechanisms of LLM agents. We abbreviate the learning mechanism subclasses using initials.} 
\label{table:example}
\resizebox{\textwidth}{!}{%
\begin{tabular}{ccccccccccc} 
\toprule
\multirow{2}{*}{\textbf{Agent}} & \multirow{2}{*}{\textbf{Perception}} & \multicolumn{2}{c}{\textbf{Cognition}} & \multicolumn{2}{c}{\textbf{Memory}} & \multirow{2}{*}{\textbf{Tool}} & \multirow{2}{*}{\textbf{Environment}} & \multicolumn{3}{c}{\textbf{Learning Mechanisms}} \\ 
\cmidrule(lr){3-4} \cmidrule(lr){5-6} \cmidrule(lr){9-11} 
 &  & Model & Reasoning & Long-Term & Short-Term &  &  & LLM& Memory& Tool\\
\midrule
WebGPT~\cite{nakano2021webgpt} & Unimodal & GPT-3 & \redx & \redx & \greencheck & \greencheck & GUI & FT\&RL & \redx & \redx \\
SayCan~\cite{ahn2022SayCan} & Multimodal & PaLM & \greencheck & \redx & \greencheck & \greencheck & Physical World & ICL & \redx & \redx \\
ReAct~\cite{yao2022react} & Unimodal & PaLM & \greencheck & \redx & \greencheck & \greencheck & QA & ICL & \redx & \redx \\
Voyager~\cite{wang2023voyager} & Multimodal & GPT-4 & \greencheck & \greencheck & \greencheck & \greencheck & Game & ICL & \greencheck & \greencheck \\
Generative Agents~\cite{park2023GenerativeAgents} & Unimodal & GPT-3.5 & \greencheck & \greencheck & \greencheck & \redx & Virtual World & ICL & \greencheck & \redx \\
AppAgent~\cite{zhang2023appagent} & Multimodal & GPT-4 & \redx & \greencheck & \greencheck & \redx & GUI & ICL & \redx & \redx \\
ChemCrow~\cite{bran2023chemcrow} & Unimodal & GPT-4 & \greencheck & \redx & \greencheck & \greencheck & Chemistry & ICL & \redx & \redx \\
MemGPT~\cite{packer2023memgpt} & Unimodal & GPT-4 & \redx & \greencheck & \greencheck & \redx & QA & ICL & \greencheck & \redx \\
ChatDev~\cite{qian2024chatdev} & Unimodal & GPT-3.5 & \redx & \greencheck & \greencheck & \greencheck & Software & ICL & \greencheck & \redx \\
LEO~\cite{huang2023LEO} & Multimodal & Vicuna-7B & \greencheck & \redx & \greencheck & \redx & Virtual World & ICL\&FT & \redx & \redx \\
VideoAgent~\cite{fan2025videoagent} & Multimodal & GPT-4 & \greencheck & \greencheck & \greencheck & \greencheck & QA & ICL & \greencheck & \redx \\
Agent Q~\cite{putta2024agent-q} & Multimodal & LLaMA-3 & \greencheck & \redx & \greencheck & \redx & GUI & FT\&RL & \redx & \redx \\
Reflexion~\cite{shinn2024reflexion} & Unimodal & GPT-3 & \greencheck & \greencheck & \greencheck & \greencheck & Code Execution & ICL\&RL & \greencheck & \redx \\
HuggingGPT~\cite{shen2024hugginggpt} & Multimodal & GPT-4 & \greencheck & \redx & \greencheck & \greencheck & QA & ICL & \redx & \greencheck \\
Jarvis-1~\cite{wang2024jarvis1} & Multimodal & GPT-4 & \greencheck & \greencheck & \greencheck & \greencheck & Game & ICL & \greencheck & \greencheck \\
SWE-agent~\cite{yang2024swe} & Unimodal & GPT-4 & \greencheck & \redx & \greencheck & \greencheck & Software & ICL & \redx & \redx \\
DigiRL~\cite{bai2024digirl} & Multimodal & DigiRL-1.3B & \redx & \redx & \greencheck & \greencheck & GUI & FT\&RL & \redx & \redx \\
CLOVA~\cite{zhi2024CLOVA} & Multimodal & GPT-4 & \greencheck & \redx & \greencheck & \greencheck & QA & ICL\&FT & \redx & \greencheck \\
MDagents~\cite{kim2024MDagents} & Multimodal & GPT-4 & \greencheck & \greencheck & \greencheck & \redx & QA & ICL & \greencheck & \redx \\
OS-Copilot~\cite{wu2024os-copilot} & Multimodal & GPT4 & \greencheck & \greencheck & \greencheck & \greencheck & GUI & ICL\&FT & \greencheck & \greencheck \\
\bottomrule
\end{tabular}%
}
\end{table*}

In summary, inspired by the von Neumann architecture, we designed the corresponding LLM agent structure while conducting a comparative analysis. We also find that this structure effectively encompasses existing work, as shown in \cref{table:example}. We can observe the rise of multimodal agents, most of which possess reasoning capabilities, yet long-term memory is not widely utilized. This indicates that the framework we designed effectively summarizes the current development trends. This analogy-based design confirms the significant similarities between the two designs, inspiring us to draw insights from computer systems.

\section{Future Directions Inspired by Computers}
\label{sec:section3}
To guide future research, it is important to address key challenges and realize the future direction. Building on detailed analyses of various components of LLM agents and the comparison drawn with the von Neumann architecture, we observe striking similarities that can inspire agent design. At the same time, viewing from a computer perspective allows us to identify shortcomings in current agent designs and gain further inspiration. In this section, we outline some future directions and critical challenges that demand attention. These challenges highlight the opportunities to advance the field.

\subsection{The Principles of Building Agents}
\begin{figure}[htbp]
    \centering
    \includegraphics[width=0.45\textwidth]{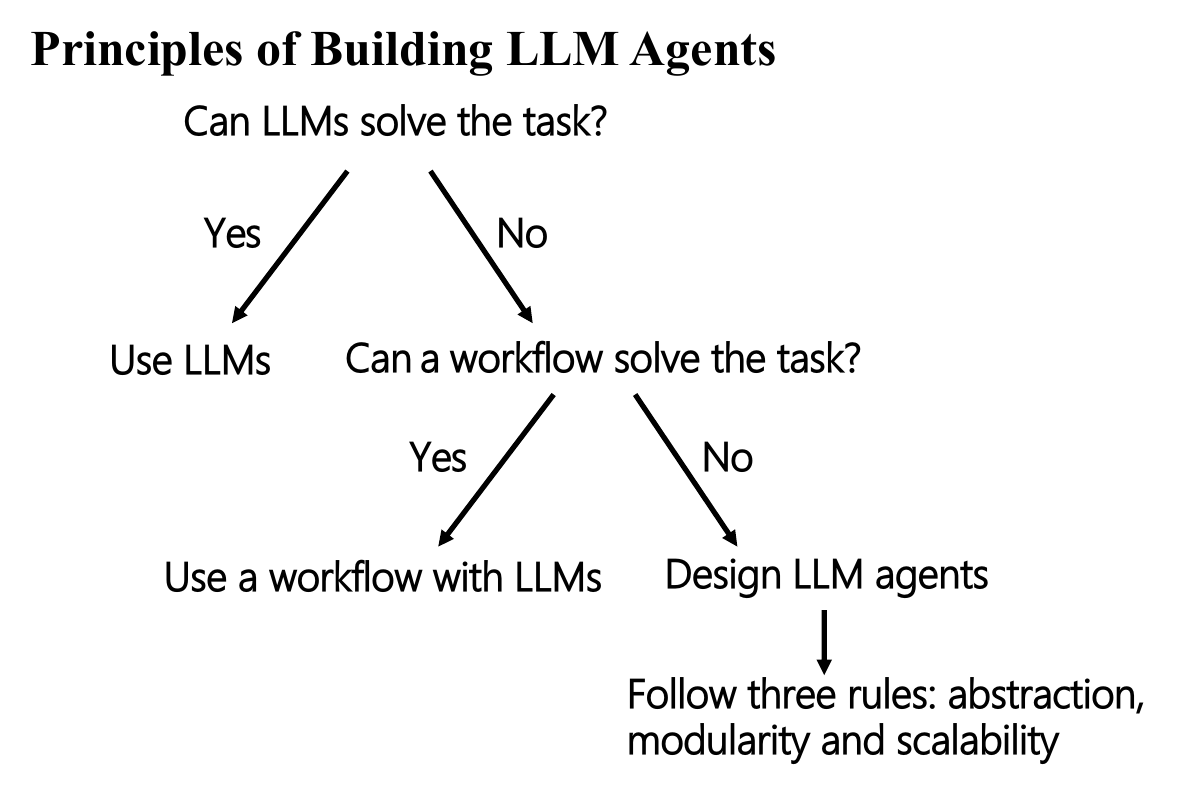}
    \caption{A design decision diagram illustrating the principles of building LLM agents.}
    \label{fig:principles}
\end{figure}

Numerous principles have been accumulated in the evolution of computers (\cref{Appdenix:C}), which guide the development of computer architectures~\cite{saltzer2009principles}. By drawing an analogy and comparing computer system design principles, we can see that building LLM agents also requires certain principles. Here, we advocate the principle of \textbf{``Build agents for demand"}. This means that agents should be designed only when a problem cannot be directly solved by an LLM itself or by constructing a workflow. For example, when an AI is needed to autonomously use various tools and browse the internet to complete tasks, agents should be built. This principle is accompanied by the adherence to three design principles: \textbf{(1)} Abstraction: Hide implementation details and provide simplified interfaces to reduce complexity. \textbf{(2)} Modularity: Break agents into functional modules with clear interfaces. \textbf{(3)} Scalability: Agents should be designed to scale efficiently with growing users, data, or computational demands without major changes. A design decision diagram is shown in \cref{fig:principles}. Moreover, we believe that more principles will be discovered.

\subsection{Refining Memory for Greater Impact}
The importance and role of memory are well recognized. In computer systems, memory is structured into finer-grained levels, incorporating cache mechanisms and diverse read-and-write operations to optimize performance and efficiency. However, such explorations remain insufficient in the domain of LLM agents, highlighting the need to refine their memory mechanisms for greater impact.

\textbf{Finer-Grained Memory}: As shown in \cref{fig:hierarchy}, computer systems form a fine-grained memory hierarchy, including components like main memory, cache, and registers. Agents can adopt this idea to design analogous modules. The existing context window is similar to main memory, and databases are akin to disks. We have found that the cache module is missing in agents. Therefore, \textbf{we call for research into the cache module in LLM agents}. Commonly used information in agents can be stored in a cache-like module, enabling quick retrieval of past experiences to efficiently complete complex tasks. This cache module not only reduces memory pressure but also facilitates the training of LLM agents. Furthermore, the construction of the memory pyramid is largely driven by the principle of locality, which emphasizes focusing on the "most important 20\%." We posit that a similar form of locality exists during the interactions of LLM agents, which could serve as a valuable guide for future memory design.

\textbf{DMA is Useful in Memory}: When drawing an analogy with computer memory mechanisms, we found that Direct Memory Access (DMA)~\cite{khawaja2014DMA} can be effective for agents. DMA improves data transfer efficiency by enabling peripheral devices (such as hard drives) to exchange data with memory directly, bypassing the CPU and enhancing overall system performance. For LLM agents, this means that access instructions do not always have to go through the LLM. This approach can optimize data retrieval by allowing long-term memory to be written directly to short-term memory. At the same time, LLMs can focus on handling other tasks more efficiently. For example, VideoAgent~\cite{fan2025videoagent} employs some unique memory querying tools to retrieve long-term memory during the question-answering process without requiring direct LLM access, enabling it to perform effectively in long-horizon video understanding.

\textbf{Memory as the Center of Data Flow}: Throughout computer system development, the von Neumann architecture initially centered on the CPU but shifted to a memory-centric model to optimize data storage and retrieval efficiency. In current agent designs, LLMs drive the agent’s reasoning processes. However, when agents interact in real-world environments, the limitations of the context window within models restrict their ability to respond effectively to rapidly changing conditions. This raises the possibility that future agent systems may adopt a memory-centric approach to handle vast external information and enable real-time interaction more effectively. Existing works, such as Generative Agents~\cite{park2023GenerativeAgents}, have explored designs in this direction, but real-time interaction remains an area requiring further study. 

\subsection{Multi-Core System}
Multi-core mechanisms in computer systems enable parallel processing by integrating multiple processing cores within a single CPU, allowing concurrent execution of tasks to enhance performance and efficiency~\cite{hennessy2011computerarch}. Each core operates independently, handling separate threads or processes, improving multitasking and throughput. This architecture optimizes resource utilization and power efficiency, making it ideal for modern computing workloads. \textbf{We advocate incorporating a multi-core mechanism into LLM agents}, utilizing an appropriate number of models to serve as the cognition module within the agent. Specifically, the unique Big.LITTLE~\cite{arm2011biglittle} architecture in multi-core systems refers to a design where high-performance "big" cores and energy-efficient "little" cores are used together, enabling optimal performance and power consumption. LLM agents can draw inspiration from this design. Larger models can be responsible for complex planning and reasoning tasks, while smaller models can manage basic operations such as dialogue and interactions with memory and tool modules. This design is analogous to the human brain's System 1 and System 2, reflecting the rationality of such a design.

\subsection{Parallelization and Pipelining}
\textbf{Parallelization}: Parallelization in computers refers to the technique of executing multiple instructions simultaneously by utilizing multiple processing threads. This highlights the correctness of breaking down large tasks into sub-tasks within agents, while also suggesting that existing agents may benefit from parallel processing. Considering parallelization from both single-agent and multi-agent perspectives, for a single agent, parallelization means that a task can be processed simultaneously by leveraging multiple tools or LLMs to achieve acceleration. In a multi-agent system, a task can be distributed among different agents, which then integrate their outputs to achieve parallel processing and improve accuracy.

\textbf{Pipelining}: Pipelining in computers is a technique that improves instruction throughput by overlapping the execution of multiple instructions. It divides the instruction cycle into discrete stages, where different stages of multiple instructions can be processed simultaneously. This allows for continuous data flow and minimizes idle time, enhancing overall performance. A similar pipelining approach can also be adopted in open-world agents to improve the speed and accuracy of processing continuous external environmental feedback, which is particularly evident in embodied intelligence and autonomous driving.

In addition to the mechanisms mentioned above, we believe that many insights can be drawn from computer systems. These directions further illustrate the strong correlation between the two fields, suggesting that the current design of LLM agents can fully leverage advancements in computer systems. Furthermore, the software layers developed through the evolution of computing—such as software systems and user interfaces—can also be applied to the advancement of LLM agents, highlighting their potential.

\section{Go beyond Computers: Learning Capability in LLM Agents}
\label{sec:section4}
If LLM agents only include the aforementioned modules and interaction mechanisms, their roles in an open-world setting would resemble that of a computer, which cannot achieve Artificial General Intelligence (AGI). To answer how general agents will evolve after construction, we find that the critical factor lies in the learning mechanism, which does not exist in computer systems. Therefore, in this section, we will explore the future learning methods of agents to go beyond computers. The analysis and summary of learning approaches in existing studies can also be found in \cref{table:example}.

\textbf{LLM Learning}: LLMs are at the heart of LLM agents, and most agent learning is achieved through the learning capabilities of LLMs. Key learning methods include In-context Learning (ICL), Fine-tuning (FT), and Reinforcement Learning (RL). In-context learning is a paradigm that allows language models to learn tasks given only a few examples in the form of demonstration~\cite{dong2022SurveyofICL}. Specifically, early studies such as ReAct~\cite{yao2022react} and Voyager~\cite{wang2023voyager} use ICL to drive LLMs in building agents, achieving good results. However, since LLMs are not specifically trained for agent tasks and in-context learning does not modify model parameters, the performance improvements achievable through in-context learning remain limited. Fine-tuning is a technique that updates model parameters on a given dataset to enable an agent to learn specific tasks. Unlike fine-tuning for LLMs, agent fine-tuning involves training on diverse levels of agent trajectory data. Consequently, many studies~\cite{chen2023fireact,zeng2023agenttuning,yin2024Lumos} focus on obtaining high-quality trajectory operation data, aiming to equip LLM agents with robust capabilities in tool usage, memory retrieval, and long-term trajectory planning. However, it cannot address out-of-domain challenges due to the absence of environmental learning. To address this issue, recent studies have attempted to integrate RL into the learning process of LLM agents, enabling them to interact with their environment. For example, AGILE~\cite{AGILE} treats the construction of LLM agents as an RL problem, using the LLM as a policy model and fine-tuning it through Proximal Policy Optimization (PPO)~\cite{schulman2017ppo}. DigiRL~\cite{bai2024digirl} trains device control agents via a two-stage fine-tuning process: first initializing the model with offline RL and then transitioning from offline to online RL. However, RL training in LLM agents often suffers from instability and the challenge of designing effective reward functions, which require further research to address. It is worth noting that these learning methods are not isolated. The practice has shown that an LLM agent often employs multiple methods simultaneously during the learning process. \textbf{This highlights that the integration of the aforementioned three approaches will likely form a standardized learning process for agents in the future.}

\textbf{Memory Manage and Tool Update}: In addition, updating the memory and tool modules is another promising direction, as such updates can enable agents to develop unique characteristics. Memory management refers to the processes of storing and organizing information during such interactions, forming a memory module characterized by specific environmental features. For instance, Generative Agents~\cite{park2023GenerativeAgents} place a memory stream at their core, extracting observations from the environment into the main memory and retrieving them based on specific rules. Without memory management, agents would lack consistent behaviors, underscoring the effectiveness of learning with memory management mechanisms. ICAL~\cite{sarch2024ICAL} introduces a method for building a memory of multimodal experiences from sub-optimal demonstrations and human feedback. It demonstrates that by distilling and filtering experiences, as the agent’s library of examples grows, the agent becomes more efficient. Tool update refers to the process by which tools are updated through interaction. The motivation for tool updates arises from the fact that some tools often become outdated or malfunction, such as an agent failing to retrieve updated data from an outdated API endpoint. One interesting work named CLOVA~\cite{zhi2024CLOVA} introduces a novel training-validation prompt tuning scheme to efficiently update tools while avoiding catastrophic forgetting. This work demonstrates that tool learning can optimize both the model and the tools themselves. In summary, \textbf{the learnable components of LLM agents are not limited to LLMs}; both the memory and tool modules can be updated and learned through interactions.
\vspace{-3pt}
\section{Conclusion}
This paper advocates for a novel perspective on building LLM agents by drawing insights from computer systems, particularly the von Neumann architecture. To support our claim, we have drawn inspiration from the von Neumann architecture and designed a structured framework consisting of distinct modules for agent components—perception, cognition, memory, tools, and actions. We then summarize future directions by drawing analogies to the development of computer systems. Key areas for improvement include establishing principles for building LLM agents, refining memory structures, utilizing multi-core systems, and leveraging parallelization and pipelining to boost LLM agent efficiency. Finally, we have investigated the learning mechanisms of LLM agents and pointed out that improved learning methods are key to their future evolution. By leveraging these insights, we aim to guide the systematic construction and evolution of general LLM agents. In addition, these insights still require thorough experimental validation, which we will continue to explore in our future research.
\nocite{langley00}

\bibliography{reference_header,references}
\bibliographystyle{icml2025}

\newpage
\appendix
\onecolumn

\section{Analogy Examples}
\label{Appdenix:A}
Our \cref{fig:teaser} compares von Neumann architecture and LLM agent architecture with task execution workflows. The structural comparison on the left side demonstrates the similarity between the two architectures, showing that they are highly alike in terms of modules and their interconnections. On the right side, the workflow comparison during task execution illustrates that they also share similarities in how tasks are performed.
\subsection{von Neumann architecture}
To intuitively demonstrate the working process of a computer (von Neumann architecture), the example in our diagram is presented using the following code. The following Python script named run.py calculates the total size of all image files in a given directory.
\begin{verbatim}
import os, sys
def Calculate_Sizes(directory):
    image_extensions = {".jpg", ".jpeg", ".png", ".gif", ".bmp", ".tiff", ".webp"}
    total_size = 0
    for root, _, files in os.walk(directory):
        for file in files:
            if os.path.splitext(file)[1].lower() in image_extensions:
                file_size = os.path.getsize(os.path.join(root, file))
                total_size = total_size + file_size
    print(f"Total size of images in '{directory}': {total_size} bytes")

if __name__ == "__main__":
    directory = sys.argv[1]
    Calculate_Sizes(directory)
\end{verbatim}

\subsection{LLM agent architecture}
For LLM agent architecture, we ask some agentic systems like ChatGPT. The query is \textbf{"Which electronic product do I use and check its latest price"}. Since the diagram we use is a preliminary conceptual framework mainly intended to illustrate the similarity between the two. Additionally, to protect personal privacy, we do not provide execution diagrams for the examples here. You can try this prompt using the most advanced model on your personal ChatGPT.

\section{Framework Formalization}
\label{Appdenix:B}

In this appendix, we provide a detailed description of each component in the proposed framework \( F = (P, C, M, T, A) \), as illustrated in ~\cref{fig:main}. This framework consists of perception, cognition, memory, tool usage, and action execution, facilitating interaction with an open-world environment.

\subsection{Perception Module}
The perception module processes external inputs from the environment, supporting both unimodal and multimodal data processing:
\begin{equation}
    P: \mathcal{O} \rightarrow \mathcal{X},
\end{equation}

where \( \mathcal{O} \) represents raw observations, and \( \mathcal{X} \) denotes the extracted feature representations. This module enables the agent to interpret sensory data from multiple sources such as images, text, and audio. It is noteworthy that past observations are often combined with past actions for perception, where actions also serve as an external source of information. Of course, in different designs, past actions and observations can be stored in memory, but this transformation does not bring about significant changes.

\subsection{Cognition Module}
The cognition module serves as the core reasoning engine, responsible for planning and decision-making. It takes perceptual inputs \( x_t \), memory content \( M_r \), and tool use output \( T_c \) as inputs to generate intermediate decisions.
\begin{equation}
    C: (\mathcal{X}, \mathcal{M}, \mathcal{T}) \rightarrow \mathcal{D},
\end{equation}
where \( \mathcal{D} \) represents the cognitive decision output space. Advanced reasoning techniques such as Chain-of-Thought (CoT), Tree-of-Thought (ToT), and Reflection mechanisms are employed to enhance decision-making.

\subsection{Memory Module}
Memory management in the framework includes both short-term (context window) and long-term (retrieval-augmented generation, RAG) storage, formulated as:
\begin{equation}
    M: \mathcal{H} \rightarrow \mathcal{M},
\end{equation}
where \( \mathcal{H} \) represents the historical interaction sequence, and \( \mathcal{M} \) denotes the memory content retrieved for current processing. Key memory operations include:
\begin{itemize}
    \item \textbf{Memory Write:} Storing new experience \( m_t \).
    \item \textbf{Memory Read:} Retrieving relevant historical data.
    \item \textbf{Memory Manage:} Optimizing memory utilization.
\end{itemize}

\subsection{Tool Module}
The tool module enables interaction with external toolset sources such as Google Search and Arxiv. It includes operations like retrieval, calling, and updating:
\begin{equation}
    T: (\mathcal{Q}, \mathcal{E}) \rightarrow \mathcal{T},
\end{equation}
where \( \mathcal{Q} \) is the query space, \( \mathcal{E} \) denotes tool set resources, and \( \mathcal{T} \) represents the tool output used in decision-making.

\subsection{Action Module}
The action module selects and executes actions based on cognitive outputs and tool assistance:
\begin{equation}
    a_t = A(C(x_t, M_r, T_c)),
\end{equation}
for time step \( t \), it has:
\begin{equation}
a_t = A\biggl( C\bigl( P(o_1, a_1, \dots, o_{t-1}, a_{t-1},o_t), M_{r}, T_{c} \bigr)\biggr),
\end{equation}
where \( T_c \) is the retrieved tool information. Actions can be categorized into internal and external types. For external actions, they alter the external environment, transitioning from \( o_{t-1} \) to \( o_{t} \). Internal actions involve searching memory, invoking tools, and performing reasoning, which do not directly affect external observations but instead modify the internal state of the LLM agents.

\subsection{Learning Paradigms}
Our framework supports multiple learning approaches to enhance performance over time:

\subsubsection{In-Context Learning}
In-context learning allows the model to make predictions based on provided examples without updating parameters:

\begin{equation}
    a_t = \arg\max_{a \in \mathcal{A}} P(a \mid o_1, a_1, \dots, o_{t-1}, a_{t-1}, o_t)
\end{equation}

\subsubsection{Fine-tuning}
Fine-tuning adjusts model parameters using a dataset \( \mathcal{D} \) with gradient-based optimization:
\begin{equation}
    \theta^* = \arg\min_{\theta} \sum_{(x, y) \in \mathcal{D}} \mathcal{L}(f_{\theta}(x), y)
\end{equation}
where \( \mathcal{L} \) represents the loss function and \( \theta \) are model parameters.

\subsubsection{Reinforcement Learning}
In reinforcement learning (RL), the agent learns through interactions with the environment by maximizing cumulative rewards:
\begin{equation}
    \pi^* = \arg\max_{\pi} \mathbb{E} \left[ \sum_{t=0}^{T} \gamma^t R(s_t, a_t) \right]
\end{equation}
where \( \pi \) is the policy, \( \gamma \) is the discount factor, and \( R \) represents the reward function.

\section{Golden insights of computer system design}
\label{Appdenix:C}
\tcbset{
  myboxstyle/.style={
    colframe=black!70,      
    colback=gray!10,        
    coltitle=white,         
    colbacktitle=black!70,  
    fonttitle=\bfseries,    
    rounded corners,          
    boxrule=0.5mm,          
    width=\textwidth,     
  }
}

\begin{tcolorbox}[myboxstyle, title={Golden insights of computer system design}]
\textbf{Abstraction}: Hide implementation details and provide simplified interfaces to reduce complexity.

\textbf{Modularity}: Break the system into functional modules with clear interfaces.

\textbf{Unified data representation}: Unified data representation ensures seamless module integration.

\textbf{Scalability}: Systems should be designed to handle growth in terms of users, data, or computational requirements without significant re-architecture.

\textbf{The end-to-end principle}: This principle states that certain functions in a system should be implemented at the endpoints rather than in the middle of a communication system. For example, reliability mechanisms like error checking are more effective when placed at the application's endpoints rather than in intermediate network layers.

\textbf{The principle of least privilege}: A system component or user should only have the minimum privileges necessary to perform its tasks. This minimizes the risk of misuse or errors affecting the overall system.

\textbf{Layering}: Systems should be built in layers, where each layer provides a specific set of services to the layer above while using the services of the layer below. This hierarchical approach aids in abstraction and simplifies debugging and maintenance.

\textbf{The robustness principle (Postel's Law)}: “Be conservative in what you do, be liberal in what you accept from others.” This principle encourages designing systems to be strict in output and tolerant in input to promote interoperability.

\textbf{Fail-Fast Systems}: A system should detect and report errors as early as possible, rather than allowing them to propagate unnoticed. This helps maintain system integrity and simplifies debugging.

\textbf{Concurrency}: Systems should efficiently manage multiple simultaneous activities, taking advantage of parallelism when possible.

\textbf{Transparency}: A system should aim to hide complexity from users where appropriate, such as in distributed systems, where failures or location details are often masked.

\textbf{Trade-off}: System design involves balancing trade-offs, such as performance vs. correctness, simplicity vs. flexibility, and latency vs. throughput.
\end{tcolorbox}

\newcommand{\StateSpace}{\mathcal{S}}
\newcommand{\ActionSpace}{\mathcal{A}}
\newcommand{\ObservationSpace}{\mathcal{O}}
\newcommand{\TransitionFunction}{T}
\newcommand{\RewardFunction}{R}
\newcommand{\ObservationFunction}{\Omega}
\newcommand{\Policy}{\pi}
\newcommand{\BeliefState}{b}

\end{document}